\documentclass{article}
\usepackage{spconf,amsmath,graphicx,amsfonts}
\usepackage{xcolor}
\usepackage{tabulary}
\usepackage{tabularx}
\usepackage{multirow}
\usepackage{subfig}
\usepackage[export]{adjustbox}

\graphicspath{{images/}}

\newcolumntype{K}[1]{>{\centering\arraybackslash}p{#1}}
\DeclareMathOperator*{\argmin}{arg\,min}

\def\x{{\mathbf x}}
\def\y{{\mathbf y}}

\def\P{\mathbb{P}}
\def\W{\mathbb{W}}
\def\F{\mathbb{F}}

\def\thetab{\boldsymbol\theta}

\title{Robustification of deep net classifiers by key based diversified aggregation with pre-filtering}
\name{Olga Taran, Shideh Rezaeifar, Taras Holotyak, Slava Voloshynovskiy\thanks{S. Voloshynovskiy is a corresponding author. The work was supported by the SNF project No. 200021\_182063.}}
\address{Department of Computer Science, University of Geneva, 7 Route de Drize, Carouge GE, Switzerland \\ \{olga.taran, shideh.rezaeifar, taras.holotyak, svolos\}@unige.ch}
\begin{document}
%
\maketitle
\begin{abstract}
In this paper, we address a problem of machine learning system vulnerability to adversarial attacks. We propose and investigate a Key based Diversified Aggregation (KDA) mechanism as a defense strategy. The KDA assumes that the attacker (i) knows the architecture of classifier and the used defense strategy, (ii) has an access to the training data set but (iii) does not know the secret key. The robustness of the system is achieved by a specially designed key based randomization. The proposed randomization 
prevents the gradients' back propagation or the creating of a "bypass" system. The randomization is performed simultaneously in several channels and a multi-channel aggregation stabilizes the results of randomization by aggregating soft outputs from each classifier in multi-channel system. The performed experimental evaluation demonstrates a high robustness and universality of the KDA against the most efficient gradient based attacks like those proposed by N. Carlini and D. Wagner \cite{carlini2017towards} and the non-gradient based sparse adversarial perturbations like OnePixel attacks \cite{su2019one}.

\end{abstract}
%
\section{Introduction}
\label{sec:intro}

Deep Neural Networks (DNN) are used to solve a wide range of problems including the classification tasks. Despite the outstanding performance and remarkable achievements, the DNN systems have recently been shown to be vulnerable to \textit{adversarial attacks} \cite{goodfellow6572explaining}. The adversarial attacks aim at tricking a decision of DNN classifiers by introducing carefully designed perturbations to a chosen target image. These perturbations, being usually quite small in magnitude and imperceptible, can drastically change the output of the classifier. This weakness seriously questions the usage of the DNN based systems in many security- and trust-sensitive applications.

In the recent years, the number of authors reported various adversarial attacks against the DNN classifiers. The diversity of discovered attacks is quite broad but without loss of generality one can cluster all attacks into three groups \cite{das2018shield, akhtar2018threat}: (1) \textit{white-box} attacks, (2) \textit{gray-box} attacks and (3) \textit{black-box attacks}. The \textit{white-box} attacks assume that the attacker has a full access to the trained model and training data. Despite a big popularity of this group of attacks, their applicability to the real-life systems is questionable. The \textit{gray} and \textit{black-box} scenarios are more suited to the real-life applications. The \textit{gray-box} attacks assume that the attacker has certain knowledge about the trained model but there exist some secret elements or an access to the intermediate results is limited. The \textit{back-box} attacks allow the attacker only to observe the output of classifier to each input without any knowledge about used architecture or possibility to observe the internal states.

\begin{figure*}[t!]
\centering
\includegraphics[width=0.6\linewidth]{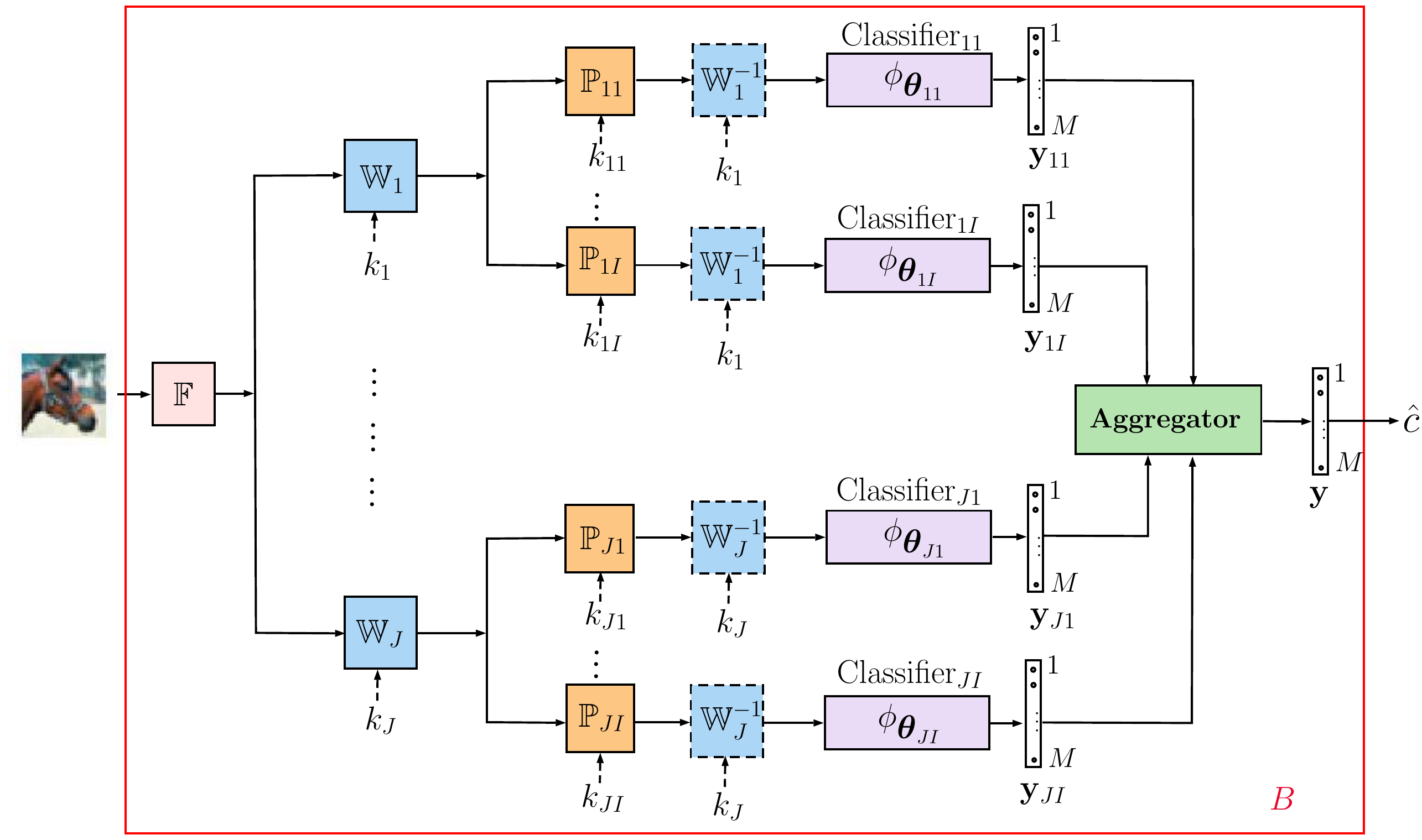}
\caption{Generalized diagram of the proposed multi-channel KDA with pre-filtering.}
\label{fig:principal classification scheme}
\end{figure*}

The existing defense mechanisms are also quite diverse \cite{Taran2018WOCM, akhtar2018threat}. However, the growing number of defenses leads to a natural invention of new and even more universal attacks. In the overwhelming majority of cases, the main interest in the adversarial attack investigation is focused on the gradient based attacks and defenses. While the non-gradient attacks and suitable defenses receive less attention but are not less dangerous and important for practice. In this respect, the goal of our paper is to investigate a new family of defense strategies that can be applied for both gradient and non-gradient based adversarial attacks in \textit{gray} and \textit{black-box} scenarios. We name it a Key based Diversified Aggregation (KDA) with pre-filtering. The generalized diagram of the proposed system is illustrated in Figure \ref{fig:principal classification scheme}. The main idea behind the KDA is to use cryptographic principles and to create an information advantage for the defender over the attacker. A secret is shared between the training and classification stages. The secret is implemented in a form of secret key used for the randomization. The system has two levels of randomization, each of which uses its own secret keys. The classification process is diversified in several channels with own randomization targeting specific randomly selected features. To reduce the negative effect of randomization, the soft outputs of multi-channels classifiers are aggregated. 

The main contribution of this paper is twofold: 
\begin{itemize}
    \item A  multi-channel classification architecture with the KDA mechanism as an universal defense strategy against the gradient and non-gradient based \textit{gray} and \textit{black-box} attacks. 
    \item An investigation of the efficiency of the proposed approach on the well-known gradient and non-gradient based adversarial attacks. 
\end{itemize}
The rest of paper is organized as follows. Section \ref{sec:milti_channel_aglo} introduces a new multi-channel classification architecture with the KDA. The efficient key-based data independent transformation is proposed in Section \ref{sec:key_based_sign_permutation}. Section \ref{sec:results} presents the empirical results obtained for the proposed algorithm. Finally, Section \ref{sec:conclusion} concludes the paper. 

\section{Classification algorithm with Key based Diversified Aggregation}
\label{sec:milti_channel_aglo}

The generalized diagram of the proposed algorithm is shown in Figure \ref{fig:principal classification scheme} and it consists of five main blocks: 

1. \textit{Pre-filtering} $\F$ that has an optional character. The goal of this block is to remove high magnitude outliers in the input images introduced by the attacker, if any. One can choose a broad range of pre-filtering algorithms from a simple local mean filter to more complex algorithms as, for example, BM3D \cite{dabov2007image} or based on DNN systems \cite{vincent2010stacked}. 

2. The input signal mapping into a \textit{transform domain} via $\W_{j}$, $1 \le j \le J$. In general, the  transform $\W_{j}$ can be any linear mapper like a random projection or belong to the family of orthonormal transformations ($\W_{j}\W^{T}_{j} = \mathbb{I}$) like DFT (discrete Fourier transform), DCT (discrete cosines transform), DWT (discrete wavelet transform), etc. Moreover, $\W_{j}$ can also be a learnable  transform or even a deep encoder. However, to avoid any key-leakage from the trained transforms, we use the data independent transform $\W_{j}$ in this paper. Thus, the transform $\W_{j}$ is generated from a secret key $k_j$. Along this line, one can also envision, for example, the DCT transform with the key defined sampling in the transform domain. We will detail below the properties of this transform.

3. \textit{Data independent processing} $\P_{ji}$, $1 \le i \le I$ serves as a defense against gradient back propagation to the direct domain. As simple examples of such kind of processing one can mention a lossy sampling $\P_{ji} \in \{0, 1\}^{l \times n}$ $l < n$ of the input signal of length $n$ as considered in \cite{chen2018secure} or a lossless permutation $\P_{ji} \in \{0, 1\}^{n \times n}$ similar to \cite{Taran2018WOCM}. The sub-block sign flipping $\P_{ji} \in \{-1, 0, +1\}^{n \times n}$ presents an additional option. It should be pointed out that to make the \textit{data independent processing} irreversible for the attacker, it is preferable to use the block $\P_{ji}$ based on a secret key $k_{ji}$.

4. \textit{Classification block} can be represented by any family of classifiers. We consider a DNN based family. 

5. \textit{Aggregation block} can be any operation ranging from a simple summation to learnable operators or special aggregation networks adapted to the data or to a particular adversarial attack. We focus on additive aggregation to demonstrate the power of a simple strategy leaving the investigation of more complex aggregations to our future work.
    
As shown in Figure \ref{fig:principal classification scheme},  in the proposed architecture the principal blocks are organized in a parallel multi-channel structure that can be followed by one or several \textit{aggregation blocks}. The final decision is made based on the aggregated result. The rejection option can  naturally be also envisioned.   

It should be pointed out that the access to the intermediate results inside the considered system provides the attacker a possibility to use the full system as a \textit{white-box}. The attacker can discover the secret keys $k_j$ and/or $k_{ji}$, make the the system end-to-end differentiable using the Backward Pass Differentiable Approximation technique \cite{athalye2018obfuscated} or via replacing the key based blocks by the bypass mappers. Therefore, it is important to restrict the access of the attacker to the intermediate results within the block $B$ (see Figure \ref{fig:principal classification scheme}). That satisfies our assumption about \textit{gray} and \textit{black-box} attacks. Additionally, it is in the accordance with the Kerckhoffs's cryptographic principle when we assume that the algorithm and architecture are known to the attacker besides the used secret key.
  
The training of the described classification architecture can be performed as follows:
\begin{equation}
\label{eq:general_formula1}
(\boldsymbol{\hat{\vartheta}}, \{\boldsymbol{\hat{\theta}_{ji}} \}) = \argmin_{\boldsymbol{\vartheta}, \{\boldsymbol{\theta}_{ji}\}} \sum_{t=1}^T \sum_{j=1}^J \sum_{i=1}^{I_j} \mathcal{L}(\y_t, A_{\boldsymbol{\vartheta}}(\phi_{{\boldsymbol\theta}_{ji}}(f(\x_t)))), \\
\end{equation}
with: 
\begin{displaymath}
\label{eq:general_formula2}
f(\x_t) = \W_j^{-1}\P_{ji}\W_j \; \F(\x_t),
\end{displaymath}
where $\mathcal{L}$ is a classification loss, $\y_t$ is a vectorized class label of the sample $\x_t$, $A_{\boldsymbol{\vartheta}}$ corresponds to the aggregation operator with parameters ${\boldsymbol{\vartheta}}$, $\phi_{\thetab_{ji}}$ is the $i$th classifier of the $j$th channel, $\thetab$ denotes the parameters of the classifier, $T$ equals to the number of training samples, $J$ is the total number of channels and $I_j$ equals to the number of classifiers per channel $j$ that we will keep fixed and equals to $I$ for all channels.

\section{Randomization using key based sign flipping in the DCT domain}
\label{sec:key_based_sign_permutation}

\begin{figure}[t!]
     \centering
     \subfloat[][sub-bands]{
        \quad\includegraphics[width=0.12\linewidth,valign=b]{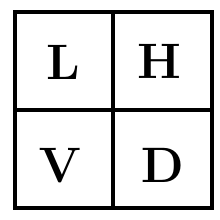}\label{fig:dct_subbands}\quad
     }
     \subfloat[][original]{
        \quad\includegraphics[width=0.125\linewidth,valign=b]{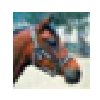}\label{fig:permut_dct_original}\quad
     }       
     \subfloat[][V]{
        \quad\includegraphics[width=0.125\linewidth,valign=b]{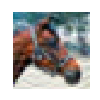}\label{fig:permut_dct_v}\quad
     }
     \subfloat[][H]{
        \quad\includegraphics[width=0.125\linewidth,valign=b]{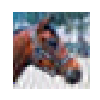}\label{fig:permut_dct_h}\quad
     }     
     \subfloat[][D]{
        \quad\includegraphics[width=0.125\linewidth,valign=b]{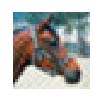}\label{fig:permut_dct_d}\quad
     }     
     \caption{Local key based sign flipping in the DCT sub-bands.}
     \label{fig:dct_sign_permut}
\end{figure}

The core element of the defense in the proposed multi-channel architecture shown in Figure \ref{fig:principal classification scheme} is a data independent processing $\P$ in a transform domain $\W$. 

In our implementation, we use the DCT as a $\W$ and the local sign flipping $\P_{ji} \in \{-1, 0, 1\}^{n \times n}$ based on the individual secret key $k_{ji}$ for each classifier. The term \textit{local} means that the processing is done only in some sub-band or block of the input signal. In general, the signal can be split into overlapping or non-overlapping sub-bands of different sizes and different positions that are kept in secret. In our experiments for the simplicity and interpretability, we split the signal in the DCT domain into four non-overlapping fixed sub-bands of the same size denoted as: (\textit{L}) top left that represents the low frequencies of the image, (\textit{V}) vertical, (\textit{H}) horizontal and (\textit{D}) diagonal sub-bands as illustrated in Figure \ref{fig:dct_subbands}. The key based sign flipping is applied independently in \textit{V}, \textit{H} and \textit{D} sub-bands keeping all other sub-bands unchanged. The effects of such processing after the inverse DCT transform are perceptually almost unnoticeable and exemplified in Figure \ref{fig:permut_dct_v} - \ref{fig:permut_dct_d}.

\begin{figure}[t!]
\centering
\includegraphics[width=1\linewidth]{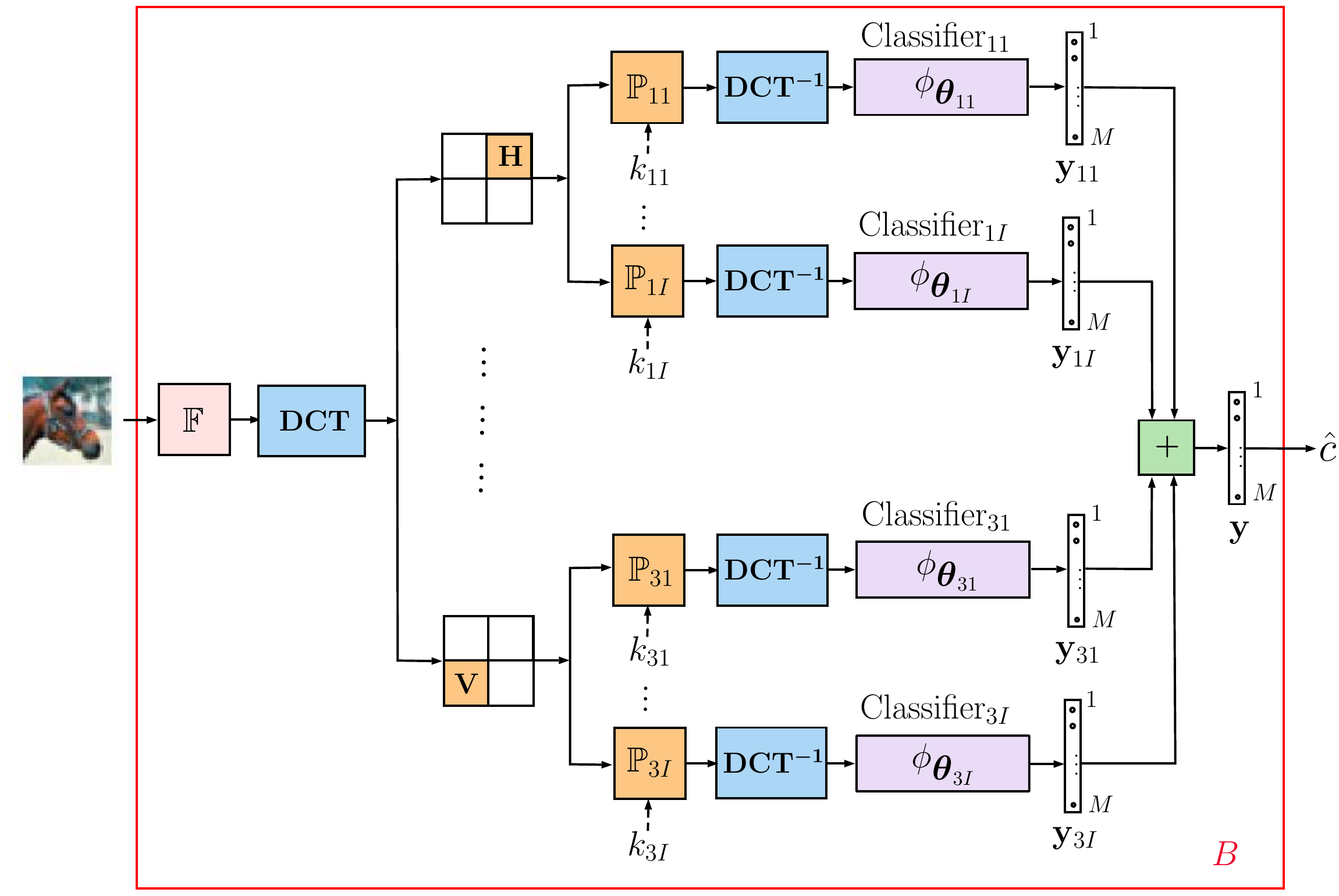}
\caption{Classification with the local DCT sign flipping.}
\label{fig:classification_via_dct_permut}
\end{figure}

The corresponding multi-channel architecture is illustrated in Figure \ref{fig:classification_via_dct_permut}. For simplicity, as an aggregation operator $A$ we use a simple summation. For the pre-filtering $\F$ we use a filter based on a difference of the point of interest in the center of the window with the median value in the window of size $3 \times 3$ around this point. If the magnitude of difference exceeds a specified threshold, the pixel is considered to be corrupted by the adversary and its value is replaced by a mean value computed in the window or otherwise, it is kept intact. Finally, each classifier $\phi_{{\boldsymbol\theta}_{ji}}$ is trained independently as: 
\begin{equation}
\boldsymbol{\hat{\theta}}_{ji} = \argmin_{\boldsymbol{\theta}_{ji}} \sum_{t=1}^T \mathcal{L}(\y_t, \phi_{{\boldsymbol\theta}_{ji}}(\W^{-1}\P_{ji}\W \; \F(\x_t))), \\
\label{eq:local_dct_sign_flipping}
\end{equation}
to ensure the best recognition in each channel under the introduced perturbation. 
%

\section{Experimental results}
\label{sec:results}

\subsection{Setup}
\label{subsec:setup}

The efficiency of the proposed multi-channel architecture diversified and randomized by the key based sign flipping in the DCT domain against the adversarial attacks was tested for two scenarios:

1. \textit{Gray-box} gradient based attack. As a gradient based attack we use the attack proposed in \cite{carlini2017towards}. This attack is among the most efficient attacks against many proposed defense strategies. Further it will be referred to as C\&W. In our experiment we use the C\&W attacks based on $\ell_2$, $\ell_0$ and $\ell_\infty$ norms.

2. \textit{Black-box} non-gradient based attack. As a non-gradient based attack we use the OnePixel attack proposed in \cite{su2019one} that uses a Differential Evolution (DE) optimisation algorithm \cite{storn1997differential} for the attack generation. The DE algorithm doesn't require the objective function to be differentiable or known but instead it observes the output of the classifier used as a black box. The OnePixel attack aims at perturbing limited number of pixels in the input image. In our experiments, we use this attack to perturb 1, 3 and 5 pixels.

For the fair comparison, the gradient based attacks were tested on the classifier with the architecture identical to those tested in \cite{carlini2017towards}. The non-gradient based attacks were tested for the classifiers based on the VGG16 \cite{simonyan2014very} an ResNet18 \cite{he2016deep} architectures used in  \cite{storn1997differential}.

All experiments have been done on the CIFAR-10 dataset \cite{krizhevsky2014cifar} that presents a particular interest as a data set with the images close to natural ones. The CIFAR-10 consists of 60000 colour images of size $32 \times 32$  (50000 train and 10000 test) with 10 classes. Due to the fact that the attack generation process is sufficiently slow for all considered attacks the experimental results were obtained on the first 1000 test samples.

\subsection{Empirical results and analysis}
\label{subsec:results}

The results obtained for the gradient based attacks in the \textit{gray-box} scenario are given in Table \ref{tab:cw_attacks_res}. The results obtained for the non-gradient based attacks in the \textit{black-box} scenario are shown in Table \ref{tab:one_pixel_attacks_res}. In both cases the column "vanilla" corresponds to the accuracy of the original classifier without any defense. The row "original" corresponds to the use of non-attacked original data.

\begin{table}[t!]
{\footnotesize
\renewcommand*{\arraystretch}{1.25}
\centering
\begin{tabular}{K{2cm}K{1.15cm}K{1.15cm}K{1.1cm}K{1.15cm}} \hline
\multirow{ 2}{*}{Attack type}  & \multirow{ 2}{*}{Vanilla} & \multicolumn{3}{c}{$J \cdot I$} \\ \cline{3-5}
& & 3 & 6 & 9 \\ \hline
Original           & 21  & 21.2  & 19.6  & 19.4  \\ 
C\&W $\ell_2$      & 100 & 22.42 & 21.3  & 21.04 \\ 
C\&W $\ell_0$      & 100 & 24.58 & 23.52 & 23.03 \\ 
C\&W $\ell_\infty$ & 100 & 22.8  & 21.39 & 21.21 \\  \hline
\end{tabular}
}
\caption{Classification error (\%) on the first 1000 test samples against the \textit{gray-box} gradient-based attacks.}
\label{tab:cw_attacks_res}
\end{table}


The analysis of the obtained results for the \textit{gray-box} gradient based attacks and the original non-attacked data demonstrates that the use of the proposed multi-channel architecture allows to improve the classification accuracy of vanilla classifier. This is  quite remarkable by itself since it shows that the multi-channel processing with the aggregation does not degrade the performance due to the introduced randomization in contrast to many defense strategies based on gradient obfuscation or detection and rejection of attacked data mechanisms.  In the case of attacked data, the C\&W attacks achieves the 100\% classification error on the vanilla undefended classifier thus showing a complete vulnerability of this deep classifier. At the same time, the use of the multi-channel architecture based on the same type of classifier with the proposed defense strategy improves the classification accuracy to the similar level of the vanilla classifier on the original non-attacked data. In the worst case of C\&W $\ell_0$ attack, the classification error is only about 2\% higher than on the original data. 

\begin{table}[t!]
{\footnotesize
\renewcommand*{\arraystretch}{1.25}
\centering
\begin{tabular}{K{2cm}K{1.15cm}K{1.15cm}K{1.1cm}K{1.15cm}} \hline
\multirow{ 2}{*}{Attack type}  & \multirow{ 2}{*}{Vanilla} & \multicolumn{3}{c}{$J \cdot I$} \\ \cline{3-5}
& & 3 & 6 & 9 \\ \hline
\multicolumn{5}{c}{VGG16}  \\ \hline
Original        & 10.7  & 11   & 9.2 & 8.9 \\ 
OnePixel $p=1$  & 58.04 & 11   & 9.5 & 8.7 \\ 
OnePixel $p=3$  & 72.13	& 10.9 & 8.9 & 8.3 \\ 
OnePixel $p=5$  & 79.02 & 12.1 & 9.3 & 9.1 \\  \hline
\multicolumn{5}{c}{ResNet18}  \\ \hline
Original        & 9.5   & 11.1 & 9.1 & 7.8 \\ 
OnePixel $p=1$  & 36.96	& 11.5 & 9   & 7.7 \\ 
OnePixel $p=3$  & 49.85 & 11.5 & 9.1 & 7.8 \\ 
OnePixel $p=5$  & 59.74	& 11.7 & 9.2 & 7.8 \\  \hline
\end{tabular}
}
\caption{Classification error (\%) on the first 1000 test samples against the \textit{black-box} non-gradient based attacks.}
\label{tab:one_pixel_attacks_res}
\end{table}
%

In the case of \textit{black-box} non-gradient based attacks, the use of the KDA improves the classification accuracy of the vanilla classifier similar to the previous case. In contrast to the \textit{gray-box} scenario, the \textit{black-box} attacks are not so harmful against the vanilla classifier. In the case of VGG16, the classification error of the vanilla classifier is about 60-80\%. In the case of ResNet18, it is about 35-60\%. For both classifiers the increase of the number of perturbed pixels ($p$) leads to the increase of classification error. The use of the proposed defense mechanism based on the KDA architecture allows to decrease the classification error to the level of classification on the original data or in other words it diminished the effect of these attacks. 

In summary, one can conclude that the obtained results indicate that the KDA  architecture with the proposed defense strategy demonstrates the high robustness to the gradient and non-gradient based attacks in the \textit{gray} and \textit{black-box} scenarios. Moreover, it allows to improve the classification accuracy of the vanilla classifiers. Finally, it should be pointed out that in all cases the increase of the number of classification channels and \textit{data independent processing} $\P_{ij}$ leads to improving the classification accuracy. However, a trade-off between the further decrease of the classification error and the increase of the complexity of the algorithm should be carefully addressed that goes beyond the scope of this paper.

%
%

\section{Conclusions}
\label{sec:conclusion}

In this paper, we considered the defense mechanism against the gradient and non-gradient based \textit{gray} and \textit{black-box} attacks. The proposed mechanism is based on the multi-channel architecture with the randomization and the aggregation of classification scores. It is remarkable that the architecture of the defense is not tailored for each class of attacks and is uniformly used for both attacks. It is also interesting to note that the diversified classification with the aggregation of the outputs of classifiers allows not only to withstand the attacks but it also improves the accuracy of vanilla classifier. It is also important to remark that the proposed approach is compliant with the cryptographic principles when the defender has an information advantage over the attacker. In our future research, we plan to extend the aggregation mechanism to more complex learnable strategies instead of used summation. 

\clearpage
\newpage
\bibliographystyle{IEEEbib}
\bibliography{strings,refs}

\end{document}